\documentclass[letterpaper]{article} %
\usepackage{aaai25}  %
\usepackage{times}  %
\usepackage{helvet}  %
\usepackage{courier}  %
\usepackage[hyphens]{url}  %
\usepackage{graphicx} %
\usepackage{natbib}  %
\usepackage{caption} %
\usepackage{algorithm}
\usepackage{algorithmic}

\usepackage{newfloat}
\usepackage{listings}
\DeclareCaptionStyle{ruled}{labelfont=normalfont,labelsep=colon,strut=off} %
\floatstyle{ruled}
\newfloat{listing}{tb}{lst}{}
\floatname{listing}{Listing}
\usepackage{color}
\usepackage{xcolor}
\usepackage{xspace}
\usepackage{amsmath}
\usepackage{amssymb}

\newcommand{\mname}{FunEditor\xspace}

\title{FunEditor: Achieving Complex Image Edits\\via Function Aggregation with Diffusion Models}
\author{
    Mohammadreza Samadi\textsuperscript{\rm 1},
        Fred X. Han\textsuperscript{\rm 1},
    Mohammad Salameh\textsuperscript{\rm 1},\\
    Hao Wu\textsuperscript{\rm 3},
    Fengyu Sun\textsuperscript{\rm 3},
    Chunhua Zhou\textsuperscript{\rm 3},
    Di Niu\textsuperscript{\rm 2}
}
\affiliations{
    \textsuperscript{\rm 1}Huawei Technologies Canada.
    \textsuperscript{\rm 2}Dept. ECE, University of Alberta.
    \textsuperscript{\rm 3}Huawei Kirin Solution, China.
    \{mohammadreza.samadi, fred.xuefei.han1, mohammad.salameh, wuhao135, zhouchunhua\}@huawei.com \\
    sunfengyu@hisilicon.com, dniu@ualberta.ca
}

\usepackage{bibentry}

\begin{document}

\maketitle

\begin{abstract}
Diffusion models have demonstrated outstanding performance in generative tasks, making them ideal candidates for image editing. Recent studies highlight their ability to apply desired edits effectively by following textual instructions, yet with two key challenges remaining. First, these models struggle to apply multiple edits simultaneously, resulting in computational inefficiencies due to their reliance on sequential processing. Second, relying on textual prompts to determine the editing region can lead to unintended alterations to the image.  We introduce \mname, an efficient diffusion model designed to learn atomic editing functions and perform complex edits by aggregating simpler functions. This approach enables complex editing tasks, such as object movement, by aggregating multiple functions and applying them simultaneously to specific areas. Our experiments demonstrate that \mname significantly outperforms recent inference-time optimization methods and fine-tuned models, either quantitatively across various metrics or through visual comparisons or both, on complex tasks like object movement and object pasting. In the meantime, with only 4 steps of inference, \mname achieves 5--24$\times$ inference speedups over existing popular methods. The code is available at: mhmdsmdi.github.io/funeditor/.
\end{abstract}

\section{Introduction}
\label{sec:intro}

Diffusion models (DM) like DALLE-3 and Adobe Firefly have significantly advanced the field of image editing, marking a paradigm shift in how digital content is generated and refined. These models have revolutionized the creative process by allowing users to produce high-quality, photorealistic images from simple textual descriptions, a capability known as text-to-image (T2I) generation. Beyond generating images, diffusion models have also excelled in following text-based instructions to perform specific editing tasks, referred to as instruction-based editing. This approach makes image manipulation more intuitive and accessible by allowing users to guide the editing process with natural language commands \cite{brooks2023instructpix2pix, sheynin2024emu, geng2024instructdiffusion}. One prominent example of an instruction-based DM is InstructPix2Pix (IP2P) \cite{brooks2023instructpix2pix}, which can edit images using textual instructions. EmuEdit \cite{sheynin2024emu} further improves the generalizability of IP2P by training on more diversified tasks. They found that training on multi-task datasets can enhance the models' editing capabilities when handling challenging instructions.

\begin{figure}[t!]
    \centering
    \includegraphics[width=\linewidth]{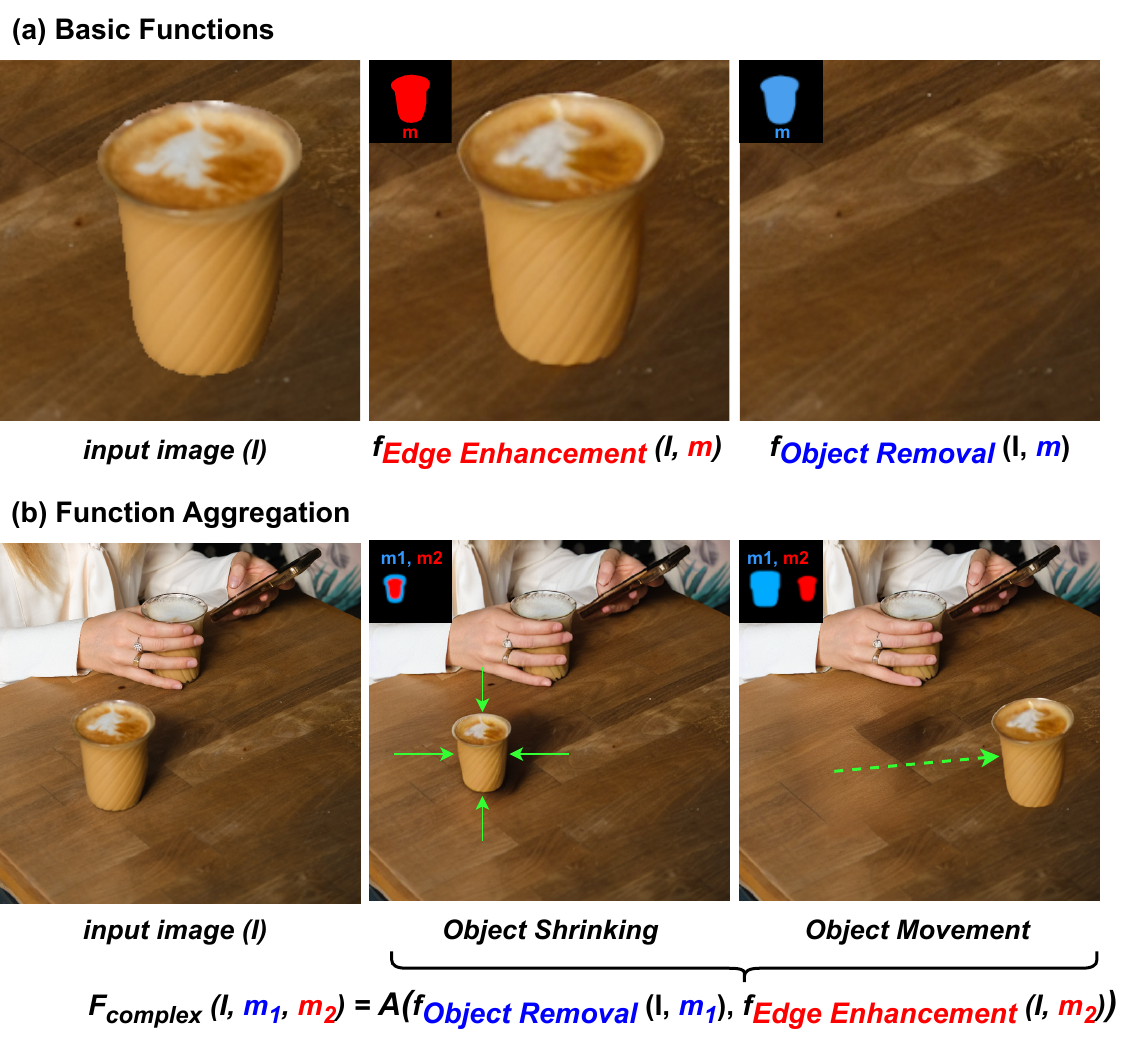}
    \caption{
        (a) Results of applying the two \textit{basic functions}—Edge Enhancement (center) and Object Removal (right)—using their respective masks on the input image Best viewed when enlarged.
        (b) Demonstration of \textit{function aggregation} using the proposed method. By simultaneously applying Object Removal and Edge Enhancement on different masks, complex edits such as object shrinking (middle) and object movement (right) can be achieved. $\mathcal{A}$ represents the operation of function aggregation.
        }
    \label{fig:edit_functions}
\end{figure}

\begin{figure*}[t!]
    \centering
    \includegraphics[width=\linewidth]{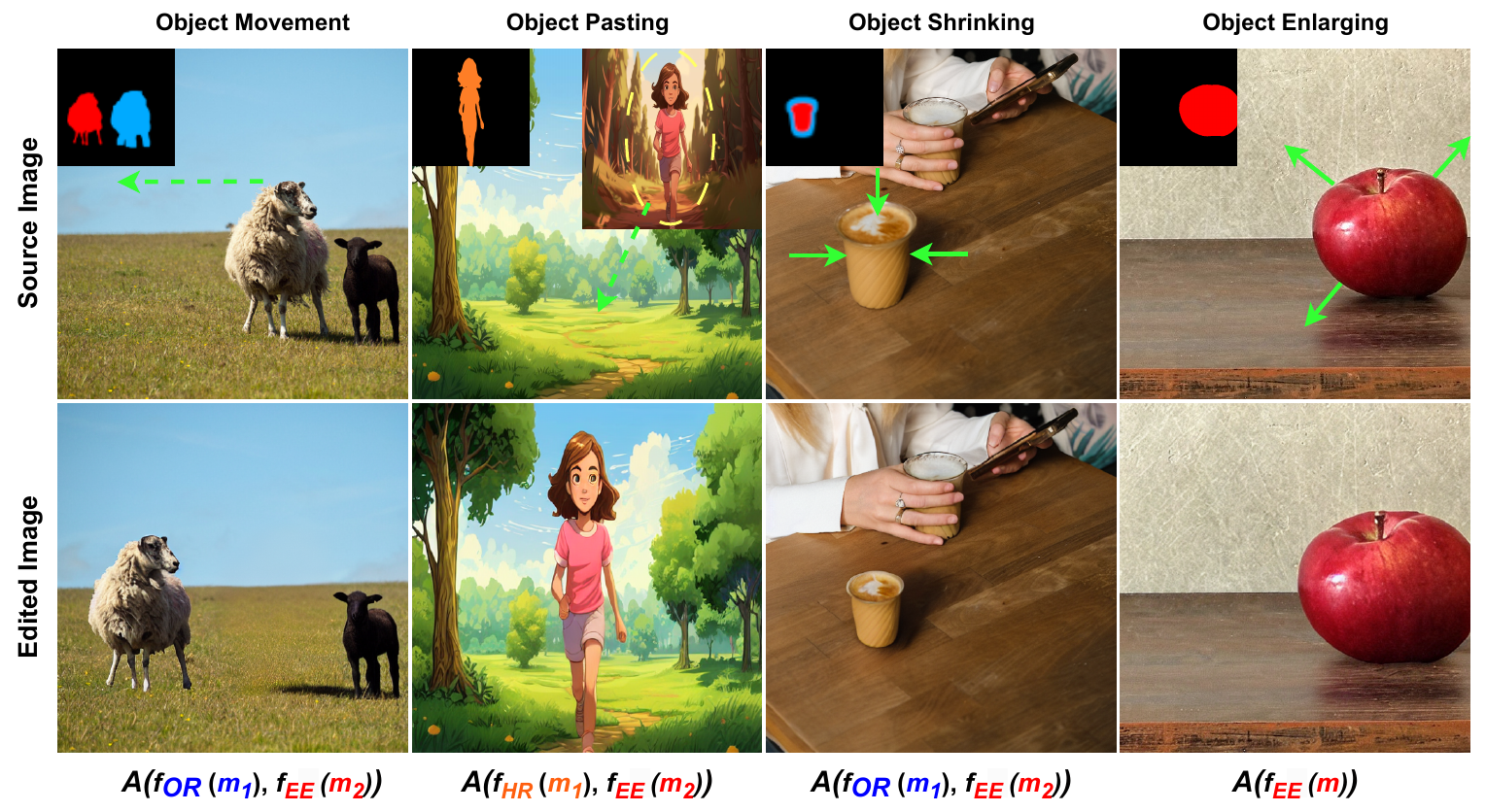}
    \caption{
        Our approach is capable of composing multiple editing functions and applying them simultaneously. This enables it to perform complex edit functions such as object movement, object resizing, and object pasting in \textbf{4 steps}. $f_{OR}$, $f_{EE}$, and $f_{HR}$ refer to object removal, edge enhancement, and harmonization functions, respectively. Each function is applied only to the specified mask region. To save space, source image I is omitted from the function arguments.
        }
    \label{fig:demo}
\end{figure*}

In recent years, the scope of instruction-based editing has expanded to include more complex tasks such as object movement \cite{wang2024repositioning, mou2024diffeditor}, subject composition \cite{yang2022paint}, and virtual try-on \cite{chen2024anydoor}. In real-world scenarios, a single edit may involve multiple types of modifications to an image, resulting in a complex edit. For instance, the object movement task involves a combination of simpler editing functions: the model needs to fill in the missing region at the original object's location with a natural background, while enhancement functions are often required for the moved object to ensure consistency with its new location. Recent studies \cite{guo2024focus} reveal that instruction-based DMs, such as IP2P, struggle to faithfully follow editing instructions when multiple edits are required on an image, highlighting potential areas for improvement. Moreover, both IP2P and EmuEdit are limited to sequential editing, where each subsequent edit is performed on the results of the previous one. This approach increases the number of denoising steps during inference and is prone to error accumulation. To address this challenge, developing methods that can efficiently handle complex editing tasks is essential.

Previous works have attempted to tackle complex tasks either through inference-time optimizations \cite{mou2024dragondiffusion, mou2024diffeditor} or supervised fine-tuning \cite{wang2024repositioning, winter2024objectdrop}. DiffEditor \cite{mou2024diffeditor} and DragonDiffusion \cite{mou2024dragondiffusion} rely on inference-time optimizations using energy-based loss functions, which involve an additional outer optimization loop and multiple calls to the UNet during inference, leading to high latency. Supervised fine-tuning approaches, such as AnyDoor \cite{chen2024anydoor}, address this issue by directly training a diffusion model capable of complex edits. However, collecting training data for all possible complex tasks is challenging, often requiring significant cost and human effort.

In this work, we propose \mname, an efficient diffusion model that performs complex edits through the aggregation of simpler functions.
Under this formulation, \mname learns to effectively integrate multiple localized edit functions and apply them simultaneously. This is achieved by defining learnable task tokens during training and combining them during inference. For localization, each simple task requires a binary mask as input and performs cross-attention modification \cite{guo2024focus, simsar2023lime}.
Compared to prior editing models, \mname offers three unique advantages.
\textbf{First}, during training, it only requires simple and atomic editing tasks such as object removal, harmonization, edge enhancement, etc., to learn new task tokens. During inference, by passing a sequence of learned task tokens into diffusion model, FunEditor can apply multiple functions simultaneously to accomplish a complex edit, requiring no energy guidance during inference. Figure \ref{fig:edit_functions} highlights the output of \mname for object movement and object resizing by aggregating object removal and edge enhancement.
\textbf{Second}, \mname is data-efficient as we only require training on each atomic task individually, for which training data is either abundant or easy to collect, making it a superior, low-cost option when compared to prior training-based approach like AnyDoor \cite{chen2024anydoor}. 
\textbf{Third}, \mname prioritizes localized editing where each simple edit function only applies to a specified region in the input image, leading to improved consistency.
\textbf{Lastly}, as \mname is compatible with pre-trained few-step diffusion models such as LCM \cite{luo2023latent}, these atomic functions can be combined with pixel-space operation like copy and pasting to perform high-quality simultaneous edits in only 4 steps without the need of subject regeneration or extra optimization steps during inference as indicated in Figure \ref{fig:demo}. 

To evaluate the effectiveness of \mname, we conducted extensive evaluations on object movement and object pasting as complex editing tasks using the COCOEE and ReS datasets. We compared \mname against a wide range of recently proposed editing baselines, including both inference-time optimization methods \cite{mou2024dragondiffusion} and fine-tuned models \cite{chen2024anydoor}. The results demonstrate that \mname significantly outperforms all baselines across various evaluation metrics, such as IQA and object-background consistency. Notably, \mname performs complex edits in just 4 steps, making it 5--24$\times$ faster than the baselines in the object movement task.  In addition to quantitative results,  visual comparisons in Figures \ref{fig:exp_movement} and \ref{fig:exp_pasting} further highlight \mname's superiority in complex editing tasks.

\section{Related Work}
\label{sec:related}

\paragraph{Image Editing with Diffusion Model.}

The success of text-to-image (T2I) diffusion models in generating high-quality images has led to their widespread adoption in image editing \cite{epstein2023diffusion, wang2024repositioning, mou2024dragondiffusion, mou2024diffeditor, alaluf2023cross, hertz2023style, jeong2024visual, patashnik2023localizing}. Training-free editing methods, such as Prompt-to-Prompt \cite{hertz2022prompt}, leverage these models by swapping cross-attention maps (CA-maps) of specific tokens during inference, enabling real-time editing without the need for fine-tuning. Plug-and-Play \cite{tumanyan2023plug} extends this approach by introducing spatial features and attention maps to guide the generation process and localize changes using masks. These methods exploit the generative capabilities of T2I diffusion models to apply edits and precisely target specific areas. Another line of work defines the editing function during inference using an energy function to update the noisy latent space \cite{mou2024diffeditor, epstein2023diffusion, shi2023dragdiffusion, mou2024dragondiffusion}. This approach guides the editing direction and samples noise to align the latent space with the editing goal, as demonstrated in DragonDiffusion \cite{mou2024dragondiffusion} and DiffEditor \cite{shi2023dragdiffusion}. However, these techniques often require additional steps, such as optimization and image inversion using methods like DDIM \cite{dhariwal2021diffusion}, which can reduce efficiency.

\paragraph{Training-based Editing Model.}

As the editing functions become more complex, the T2I diffusion model may show inconsistencies and poor performance. This is primarily because the model has not seen similar examples during training. Training-based approaches address this by using large editing datasets \cite{zhang2024magicbrush, zhang2024hive, hui2024hq}, which include pairs of source and edited images along with the corresponding edit instructions, to fine-tune a T2I diffusion model specifically for editing tasks. InstructPix2Pix (IP2P) \cite{brooks2023instructpix2pix} leverages GPT-3 \cite{brown2020language} and Prompt-to-Prompt \cite{hertz2022prompt} to generate synthetic editing datasets. To enable the source image to be passed to the UNet as an additional input, they added a new channel to the input convolutional layer, allowing the model to preserve details in unintended areas. Building on the success of InstructPix2Pix (IP2P), subsequent works, such as MagicBrush \cite{zhang2024magicbrush}, have manually annotated images to create higher-quality datasets. By utilizing the same training procedure to fine-tune their editing models, these approaches have demonstrated improved performance \cite{zhang2024hive, hui2024hq}. However, the main shortcoming of IP2P-based editing models is the occurrence of unintended changes to other parts of the source image \cite{simsar2023lime, guo2024focus}. To address this issue, FoI \cite{guo2024focus} introduced an inference technique that modulates attention and masks the sampled noise for the text-conditioned output, thereby limiting changes only to the specific area of the edit.

\paragraph{Multi-task Diffusion Model.}
Unifying multiple tasks within a single model has been an intriguing research direction in the literature. InstructDiffusion \cite{geng2024instructdiffusion} investigates the potential benefits of joint training across various tasks, including keypoint detection, segmentation, image enhancement, and editing. Their results suggest that multi-task training enhances the model's performance across these diverse tasks. Similarly, EmuEdit \cite{sheynin2024emu} introduces a unified diffusion model capable of performing multiple editing and computer vision tasks, such as object removal, global editing, and object detection. EmuEdit uses task embeddings as conditions to switch between tasks, effectively allowing the model to adapt its behavior to the specific task at hand. While these approaches show impressive results on individual tasks, they are limited by their inability to apply two tasks simultaneously, requiring multiple user interactions to achieve complex edits. In this work, we address this limitation by bridging the gap between multi-task training and the ability to perform multiple editing tasks in a single operation, enabling efficient complex editing.

\section{Method}
\label{sec:method}
\begin{figure*}[!t]
    \centering
    \includegraphics[width=\linewidth]{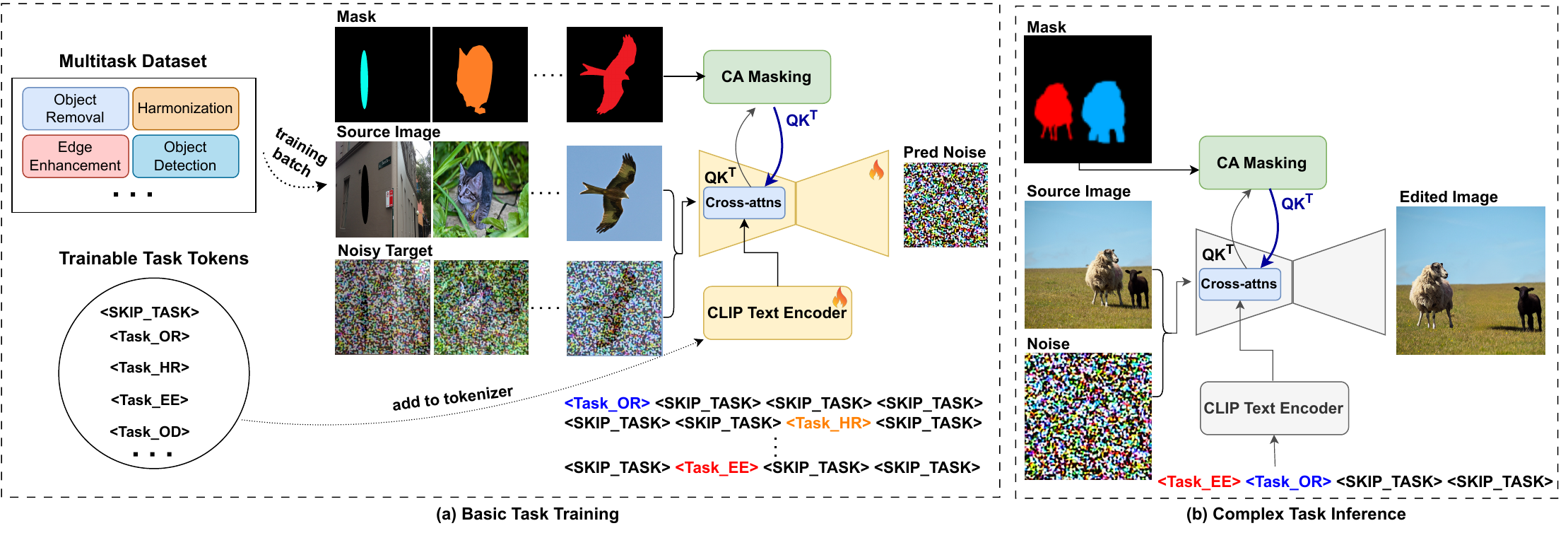}
    \caption{
        Overview of our proposed training and inference pipeline. 
        During the basic task training phase (a) the diffusion model learns to perform various simple tasks based on the provided task tokens and masks.
        During inference (b), we could implement complex edit functions by combining multiple task masks and tokens.
        }
    \label{fig:overview}
\end{figure*}

In this section, we explain how complex localized editing tasks could be formulated into a problem of function aggregation. This formulation allow us to achieve simultaneous edits during inference in few-steps. Then, we showcase how \mname solve this problem by a carefully designed training framework. Figure \ref{fig:overview} provides an overview of our proposed pipeline for both training and inference.

\subsection{Editing as Function Aggregation}
\label{sec.formulation}
Let $f$ be a localized editing function to a certain region of the source image $I$, indicated by a binary mask $m$. $f(I,m)$ denotes the edited image.
We then define an aggregation function $F$ as:

\begin{equation}
    \label{eq:functional_editing}
    F = \mathcal{A}(f_1(I, m_1), f_2(I, m_2), \dots, f_n(I, m_n)).
\end{equation}

In essence, $F$ represents aggregating different $f$ and applying them simultaneously to multiple regions of the source image in one shot.
The main advantage of such formulation is that it enables the divide-and-conquer of complex tasks (see Figure \ref{fig:edit_functions}).
Given that it is challenging to collect sufficient high-quality training data to directly learn $F$, our goal is to learn each $f$ individually during training and aggregate their effects during inference.
In \mname, each basic function $f_i$ is represented by a trainable task token $\mathbf{T}_i$.
For example, providing the token \texttt{<Task-OR>} to the UNet as input would enable the task of object removal, as indicated by Figure~\ref{fig:overview}(a).
Although EmuEdit also define task embedding vectors that are of similar purposes, the main difference is that we directly add the task tokens to the vocabulary of the prompt tokenizer.
This design offers two advantages compared to EmuEdit. 
First, it allows \mname to freely combining tasks by passing the corresponding task tokens as prompts to the UNet.
To disable tasks, we introduce another special token named \texttt{<SKIP-TASK>}, which ensures a consistent prompt length during training and inference, improving the overall generalizability.
Second, task tokens pave the way towards localized, simultaneous editing during inference.
To achieve this, we simply set the desired task tokens in the input prompt and keep the remaining ones as \texttt{<SKIP-TASK>}.
Similar to instruction-based editing \cite{simsar2023lime, guo2024focus}, our goal is to apply each editing function $f_i$ to its corresponding mask $m_i$ to maintain control over the editing region and prevent leakage into other areas. 

Since the UNet processes the task tokens via cross-attentions (CA), it is natural to inject localization information via the same medium.
Therefore, we employ a cross-attention (CA) masking technique to ensure each task $f_i$ does not affect the unmasked area $1 - m_i$. 
During both training and inference, we apply this CA masking technique to all CA layers. 
For each CA layer, we interpolate the mask into its resolution and set the values of $QK^T$ of $\mathbf{T}_i$ token corresponding to the unmasked area $1 - m_i$ to its minimum value. Mathematically represented as:
\begin{equation} 
    \label{eq:ca_masking}
    QK^T[(1 - m_i), \mathbf{T}_i] = \min(QK^T).
\end{equation}

\begin{figure}[t!]
    \centering
    \includegraphics[width=\linewidth]{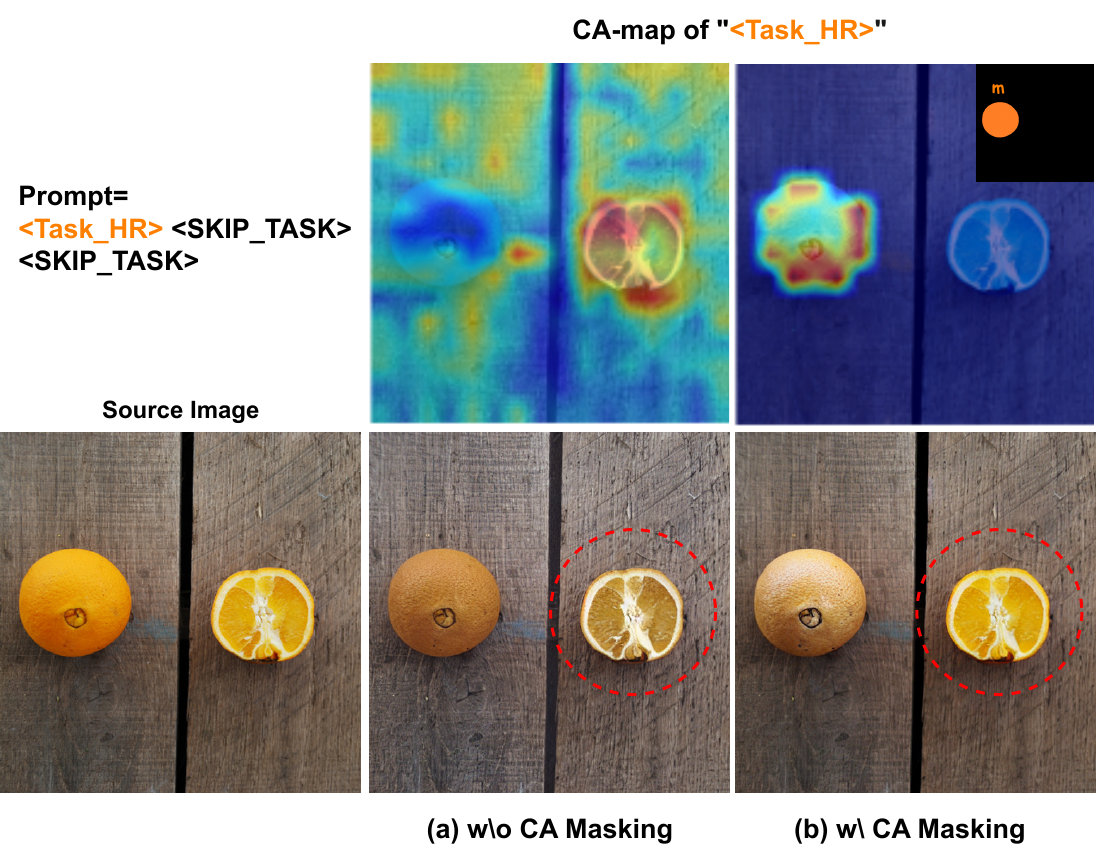}
    \caption{
        Harmonization without cross-attention masking affects the entire image (a). While with masking, edits are confined to the masked region, (b), preventing changes to unmasked areas.
        Mask is indicated by the top right mini-figure.
    }
    \label{fig:ca_masking}
\end{figure}

Figure~\ref{fig:ca_masking} highlights the importance of CA masking for localized editing. Notably without CA masking the model could attend to other regions of the source image and generates undesired changes outside of the masked region.

\begin{figure*}[!t]
    \centering
    \includegraphics[width=\linewidth]{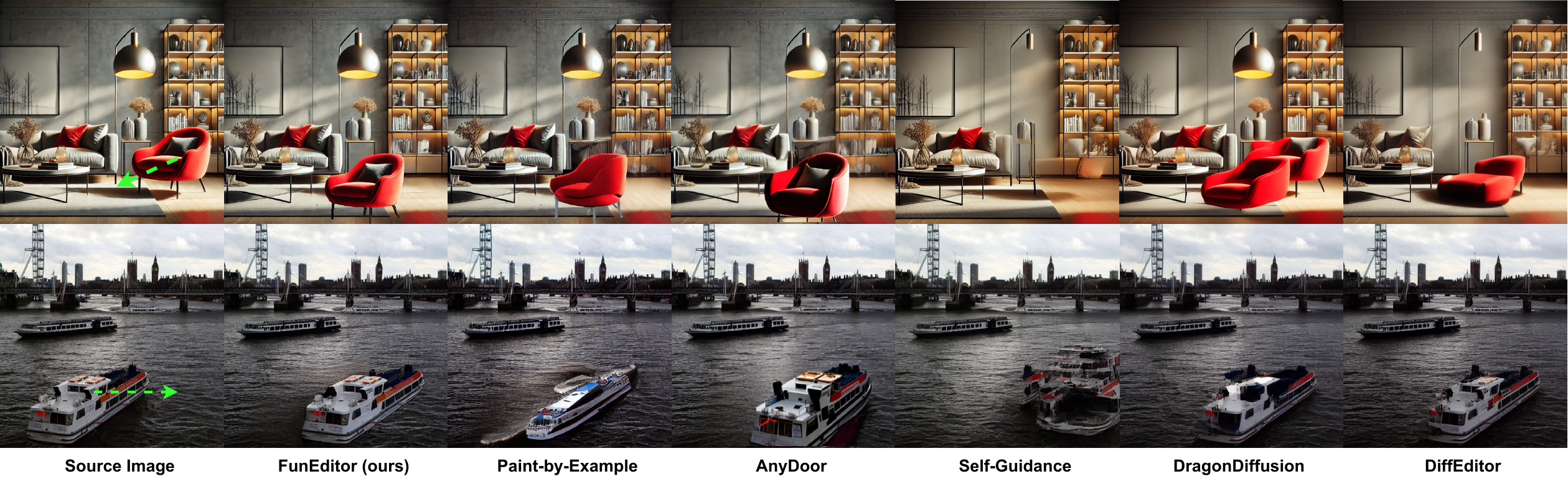}
    \caption{Qualitative comparison between our approach and baseline methods for object repositioning within an image, demonstrating the superior performance of our method. To move an object, \mname composes object removal and edge enhancement functions.}
    \label{fig:exp_movement}
\end{figure*}

\begin{figure*}[!t]
    \centering
    \includegraphics[width=\linewidth]{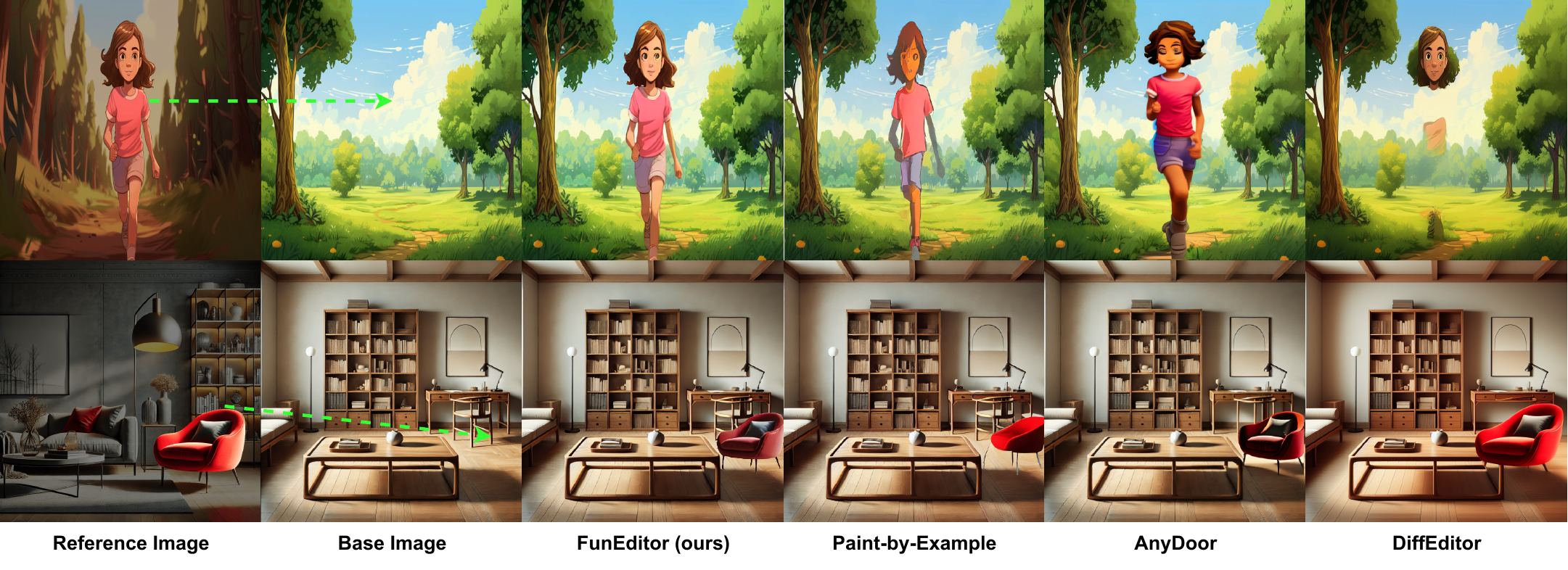}
    \caption{Visual comparison between our method and baseline methods for object pasting from a reference image into a target image. \mname applies harmonization and edge enhancement functions to seamlessly paste an object into another image.}
    \label{fig:exp_pasting}
\end{figure*}

\subsection{\mname Training Strategy}
The training procedure for \mname builds upon the IP2P architecture. In IP2P, a source image $I_{src}$, a target image $I_{trg}$, and a text instruction $T$ are provided, with noise $\epsilon$ added to $I_{trg}$ to create a noisy input $z_t$. The UNet $\epsilon_\theta$ is trained to predict $\epsilon$ using the following objective:
\begin{equation}
    \label{eq:objective_ip2p}
    \min_\theta \mathbb{E}_{y, \epsilon, t}\left[||\epsilon - \epsilon_\theta(z_t, t, I_{src}, I_{trg}, T)||^2_2\right],
\end{equation}
where $y = (I_{src}, I_{trg}, T)$ represents a sample from the IP2P editing dataset. In \mname, instead of using a text instruction $T$, we incorporate new task tokens into the CLIP text tokenizer \cite{radford2021learning} to represent the desired editing function. Additionally, we use masks corresponding to each editing function within the cross-attention. Thus, our training sample takes the form $(I_{src}, \mathbf{T}, m, I_{trg})$, where $I_{src}$ is the source image, $\mathbf{T}$ is the task token, $m$ is the mask, and $I_{trg}$ is the target image.

The idea of task embeddings provide the foundation for enabling simultaneous edits in \mname. EmuEdit defines learnable task embedding vectors, which are provided to the UNet as task-selectors via cross-attention interaction and the timestep embedding layers. In contrast, we incorporate the task embeddings via cross-attention to have more control on their spatial effects on the editing regions. The following equation defines our objective for training \mname, where we learn an embedding vector for each task token that represents a basic edit:
\begin{equation}
    \label{eq:learnable_embedding}
    \min_{\theta, v_1, \ldots, v_k} \mathbb{E}_{\hat{y}, \epsilon, t}[||\epsilon - \epsilon_\theta(z_t, t, I_{src}, m, \mathbf{T})||^2_2].
\end{equation}

Experimentally we find that setting the number of task tokens in the prompt equal to the number of supported simple tasks generates the best results.
During training we randomly activate one task token and keep the remaining tokens as \texttt{<SKIP-TASK>}.
Since the task token ordering is not important, we also randomly shuffle the task tokens in every training step to prevent the model from memorizing their positions.

\section{Experiments}
\label{sec:experiments}
In this section, we evaluate the effectiveness of our proposed method by evaluating it on to two complex image editing tasks: Object Movement ($F_{OM}$) and Object Pasting ($F_{OP}$). Building on the formulation introduced in the previous section, we define each complex function as an aggregation of the following basic functions:
\begin{align*}
    &F_{OM}(I, M_{src}, M_{trg}) = \mathcal{A}\big( f_{OR}(I, M_{src}), f_{EE}(I, M_{trg}) \big), \\
    &F_{OP}(I, M_{trg}) = \mathcal{A}\big(f_{HR}(I, M_{trg}), f_{EE}(I, M_{trg}) \big).
\end{align*}
In these formulations, $f_{OR}$, $f_{EE}$, and $f_{HR}$ represent the basic functions for object removal, edge enhancement, and harmonization, respectively. The masks $M_{src}$ and $M_{trg}$ correspond to the source object and the target location within the image.

Our method is compared to baseline approaches with respect to both quality and efficiency. The subsequent subsections will detail the experimental setup, including the datasets used and evaluation metrics. We will then present the results of our experiments, comparing the performance of our method with that of baseline techniques across the defined tasks.

\begin{table}
    \centering
    \begin{tabular}{lcccc}
    \hline
    Method & \#Steps & NFEs & Avg. Latency & std \\
    \hline
    SelfGuidance & 50 & 100 & 11 & 0.6 \\
    DragonDiffusion & 50 & 160 & 23 & 0.7 \\
    DiffEditor & 50 & 176 & 24 & 0.4 \\
    \hline
    Paint-by-Example & 50 & 50 & 5 & 0.6 \\
    AnyDoor & 50 & 50 & 12 & 0.4 \\
    \mname (ours) & 4 & 4 & 1 & 0.4 \\
    \hline
    \end{tabular}
    \caption{Efficiency is compared in terms of the number of function evaluations (NFEs) and latency (seconds). Latency is measured as the average wall-clock time for editing one image over 10 runs on a single Nvidia V100 (32GB) GPU.
    }
    \label{table:efficiency}
\end{table}

\subsection{Experimental Settings}
\subsubsection{Implementation Details}
We fine-tuned our model using 4 Nvidia V100 (32GB) GPUs, with the IP2P UNet \cite{brooks2023instructpix2pix} serving as our diffusion backbone. The AdamW optimizer \cite{loshchilov2017decoupled} was employed with a learning rate of $5 \times 10^{-5}$ and a batch size of 4. During inference, we distilled the trained model into 4 steps using SD1.5 LCM-LoRA from HuggingFace\footnote{https://huggingface.co/latent-consistency/lcm-lora-sdv1-5}. For the baselines, we used SDv1.5 \cite{Rombach_2022_CVPR} from HuggingFace\footnote{https://huggingface.co/runwayml/stable-diffusion-v1-5}, adhering to all default inference hyperparameters. As a preprocessing step for object removal, we dilated the mask with a kernel size of 20 to prevent any pixel leakage into the inpainted region.

\subsubsection{Datasets}
The COCOEE dataset, compiled by \citet{yang2022paint}, features 3,500 images manually selected from the MSCOCO (Microsoft Common Objects in Context) \cite{lin2014microsoft} validation set. A human operator used the Segment Anything model \cite{kirillov2023segment} to extract segments and assign diff vectors, resulting in a benchmark of 100 images with corresponding masks and diff vectors for object movement tasks. \citet{wang2024repositioning} also released the ReS dataset, comprising 100 pairs of real-world images that present challenging cases of object movement. The details of the training datasets for atomic edits can be found in the appendix.

\begin{table*}[!t]
    \centering
    \begin{tabular}{l|ccc|cc|cc|c}
        \hline
        Method & \multicolumn{3}{|c}{IQA} & \multicolumn{2}{|c}{Object Consistency} & \multicolumn{2}{|c}{Background Consistency} & \multicolumn{1}{|c}{Semantic Consistency} \\
         & TOPIQ $\uparrow$ & MUSIQ $\uparrow$ & LIQE $\uparrow$ & LPIPS $\downarrow$ & PSNR $\uparrow$ & LPIPS $\downarrow$ & PSNR $\uparrow$ & CLIP-I2I $\uparrow$ \\
        \hline
        SelfGuidance & 0.554 & 65.91 & 3.90 & 0.083 & 22.77 & 0.259 & 17.86 & 0.897 \\
        DragonDiffusion & 0.571 & 68.87 & 4.27 & 0.034 & 28.59 & 0.098 & 23.99 & 0.965 \\
        DiffEditor & 0.579 & 69.09 & 4.27 & 0.036 & 28.49 & 0.094 & 24.23 & 0.967 \\
        \hline
        Paint-by-Example & 0.595 & 69.57 & 4.31 & 0.066 & 23.12 & 0.089 & 25.23 & 0.937 \\
        AnyDoor & 0.558 & 68.37 & 4.12 & 0.053 & 25.43 & 0.122 & 23.55 & 0.954 \\
        \mname (ours) & \textbf{0.611} & \textbf{70.39} & \textbf{4.36} & \textbf{0.017} & \textbf{34.27} & \textbf{0.066} & \textbf{26.54} & \textbf{0.969} \\
        \hline
    \end{tabular}
    \caption{Quantitative evaluation of our approach compared to the baselines on the object movement task using the COCOEE dataset. To move an object, \mname encompasses object removal and edge enhancement functions.}
    \label{table:results}
\end{table*}

\subsubsection{Metrics}
In our evaluations, we utilize four groups of metrics: overall image quality, object consistency, background consistency, and semantic consistency. For Image Quality Assessment (IQA), we assess the perceptual quality of the images using TOPIQ \cite{chen2024topiq}, MUSIQ \cite{ke2021musiq}, and LIQE \cite{zhang2023blind}. Object Consistency is measured by comparing the similarity of the moved object to its original counterpart in the source image, using the LPIPS \cite{zhang2018unreasonable} and PSNR metrics. Background Consistency is evaluated by assessing the similarity between the background in the edited image and the original image, using the same metrics as for Object Consistency. Finally, Semantic Consistency is determined by evaluating the similarity between the semantics of the source and edited images through CLIP image encoder, which measures image-to-image similarity based on CLIP embeddings \cite{radford2021learning}.

\subsubsection{Baselines}
We evaluate our approach against both training-based and training-free methods on complex editing tasks. Given a limited number of open-source training-based model specifically for complex tasks \cite{winter2024objectdrop, wang2024repositioning}, we include training-free approaches to comprehensively assess the quantitative performance of our method. Our baselines for the complex editing tasks are as follow:
\begin{itemize}
    \item \textbf{Paint-by-Example}~\cite{yang2022paint}: This study explores a training-based approach for reference-guided image editing, training a model to fill a designated mask area with a reference object.
    \item \textbf{AnyDoor}~\cite{chen2024anydoor}: A training-based method that enables teleportation of an object into a user-specified location and shape, leveraging object identity information to accurately reconstruct the object in the intended form.
    \item \textbf{SelfGuidance}~\cite{epstein2023diffusion}: A training-free method that utilizes energy functions on attention maps to optimize the noisy latent representation during inference, allowing for edits such as object movement.
    \item \textbf{DragonDiffusion}~\cite{mou2024dragondiffusion}: A training-free approach that employs an energy function with two components: one for applying changes within the target mask and another for preserving the unmasked area.
    \item \textbf{DiffEditor}~\cite{mou2024diffeditor}: Similar to DragonDiffusion, this method enhances the quality of edits by optimizing an energy function, incorporating score-based gradient guidance and time interval strategies into the diffusion noise sampling process.
\end{itemize}

\begin{table*}[t!]
    \centering
    \begin{tabular}{l|ccc|cc|cc|c}
        \hline
        Method & \multicolumn{3}{|c}{IQA} & \multicolumn{2}{|c}{Object Consistency} & \multicolumn{2}{|c}{Background Consistency} & \multicolumn{1}{|c}{Semantic Consistency} \\
         & TOPIQ $\uparrow$ & MUSIQ $\uparrow$ & LIQE $\uparrow$ & LPIPS $\downarrow$ & PSNR $\uparrow$ & LPIPS $\downarrow$ & PSNR $\uparrow$ & CLIP-I2I $\uparrow$ \\
        \hline
        SelfGuidance & 0.586 & 69.41 & 3.61 & 0.064 & 24.21 & 0.273 & 17.92 & 0.869 \\
        DragonDiffusion & 0.690 & 74.95 & 4.72 & 0.030 & 29.68 & 0.083 & 25.38 & 0.934 \\
        DiffEditor & 0.691 & 74.94 & 4.73 & 0.032 & 29.59 & 0.083 & 25.44 & 0.933 \\
        \hline
        Paint-by-Example & \textbf{0.717} & 74.94 & \textbf{4.75} &  0.052 & 24.20 & 0.076 & 26.27 & 0.907 \\
        AnyDoor & 0.679 & 73.68 & 4.51 & 0.045 & 25.89 & 0.106 & 23.48 & 0.923  \\
        \mname (ours) & 0.702 & \textbf{75.00} & \textbf{4.75} & \textbf{0.015} & \textbf{35.61} & \textbf{0.063} & \textbf{26.48} & \textbf{0.942} \\
        \hline
    \end{tabular}
    \caption{Quantitative evaluation of our approach compared to the baselines on the object movement task was conducted using the ReS dataset. To move an object, \mname encompasses object removal and edge enhancement functions.}
    \label{table:results_Res}
\end{table*}

\subsection{Experimental Results}
\subsubsection{Efficiency}
Table \ref{table:efficiency} demonstrates that \mname, using only $4$ steps, requires $172$ fewer NFEs and is $23$ seconds faster than the state-of-the-art method for object movement, DiffEditor \cite{mou2024diffeditor}, on the COCOEE dataset. SelfGuidance, DragonDiffusion, and DiffEditor, rely on optimizing energy functions to guide complex edits during inference. These method requires multiple calls to the UNet, as it must perform both inpainting and blending of the object with its surrounding environment, leading to a more time-consuming process. Comparing to training-based approaches, our model is $12$ times faster than the training-based baseline AnyDoor \cite{chen2024anydoor} and $5$ times faster than Paint-by-Example \cite{yang2022paint}. These methods regenerate the object to fit the target position demanding a higher number of steps to achieve the desired result. In contrast, \mname achieves superior results in fewer steps by employing simpler enhancement functions, such as edge enhancement, and pixel operations, without the need for fully regenerating the object. Instead, these functions refine the composition and harmonization of the pasted object in pixel space, enabling us to achieve complex edits more efficiently. It should be noted that the studies that used synthetic data generation for object movement \cite{winter2024objectdrop, wang2024repositioning} did not open-source their models, therefore we could not use them as baselines to assess their latency in a multi-turn fashion.

\subsubsection{Image Quality} We evaluate the performance of \mname through quantitative and qualitative experiments on object movement and object pasting. Tables \ref{table:results} and \ref{table:results_Res} presents our quantitative comparison with the baselines on the object movement task on COCOEE and ReS datasets. Our model achieves higher quality across almost all IQA metrics compared to the baselines, improving the accuracy of object movement while preserving the details of both the object and the background. By aggregating atomic functions, \mname successfully removes objects, enhancing semantic consistency and overall image quality compared to optimization-based approaches that rely on energy functions. Additionally, since we use the original object in the pixel space and apply enhancement functions such as edge enhancement and harmonization, we can faithfully move or paste the object without altering its appearance, unlike the baselines that regenerate the object and modify its appearance.

Figure \ref{fig:exp_movement} and \ref{fig:exp_pasting} provide qualitative comparisons between our method and the baselines for object movement and object pasting. As shown in Figure \ref{fig:exp_movement}, the baselines face two primary issues: unsuccessful object removal and unsuccessful object pasting. In contrast, our method seamlessly repositions and blends the object into the destination while effectively removing the original object. Supporting the quantitative results in Table \ref{table:results}, \mname faithfully preserves the background, applying changes only to the intended editing area. For example, SelfGuidance and DiffEditor alter the shape of the floor lamp in the first image. In the case of object pasting, as shown in Figure \ref{fig:exp_pasting}, regenerating the object can lead to issues such as unfaithful or incomplete regeneration like the first example.

\section{Conclusion}
\label{sec:conclusion}
In this paper, we introduced \mname, an efficient editing approach that utilizes function aggregation to address complex editing tasks such as object movement. \mname offers several key advantages: it requires only publicly available datasets for atomic tasks, enables the simultaneous execution of multiple functions to achieve complex edits, and is compatible with few-step diffusion models. Our approach not only outperforms baselines specifically designed for complex tasks like object movement and object pasting in terms of quality but also offers superior efficiency. Additionally, we evaluated \mname across a wide range of numerical metrics and visual quality, comparing it against both training-based and training-free baselines.

\bibliography{aaai25}

\end{document}